\documentclass[sigconf]{acmart}
\usepackage{booktabs} 

\usepackage{booktabs} 

\usepackage{multirow}

\newcommand{\dataset}{POIReviewQA}

\setcopyright{rightsretained}

\copyrightyear{2018} 
\acmYear{2018} 
\setcopyright{acmcopyright}
\acmConference[GIR'18]{12th Workshop on Geographic Information Retrieval}{November 6, 2018}{Seattle, WA, USA}
\acmBooktitle{12th Workshop on Geographic Information Retrieval (GIR'18), November 6, 2018, Seattle, WA, USA}
\acmPrice{15.00}
\acmDOI{10.1145/3281354.3281359}
\acmISBN{978-1-4503-6034-0/18/11}

\begin{document}
\title[POIReviewQA]{{\dataset}: A Semantically Enriched \\
POI Retrieval and Question Answering Dataset}

\author{Gengchen Mai, Krzysztof Janowicz}
\affiliation{%
  \institution{STKO Lab, UCSB}
}
\email{{gengchen_mai, janowicz}@geog.ucsb.edu}

\author{Cheng He, Sumang Liu, Ni Lao}
\affiliation{%
  \institution{SayMosaic Inc.}
}
\email{{cheng.he, sumang.liu, ni.lao} @mosaix.ai}

\renewcommand{\shortauthors}{Gengchen Mai et al.}

\begin{abstract}

Many services that perform information retrieval for Points of Interest (POI) utilize a Lucene-based setup with spatial filtering.
While this type of system is easy to implement it does not make use of semantics but relies on direct word matches between a query and reviews leading to a loss in both precision and recall.
To study the challenging task of semantically enriching POIs from unstructured data in order to support open-domain search and question answering (QA), we introduce a new dataset {\dataset}\footnote{\url{http://stko.geog.ucsb.edu/poireviewqa/}}.
%
It consists of 20k questions (e.g.``is this restaurant dog friendly?'') for 1022 Yelp business types.
For each question we sampled 10 reviews, and annotated each sentence in the reviews whether it answers the question and what the corresponding answer is.
%
To test a system's ability to understand the text we adopt an information retrieval evaluation by ranking all the review sentences for a question based on the likelihood that they answer this question.
We build a Lucene-based baseline model, which achieves 77.0\% AUC and 48.8\% MAP.
A sentence embedding-based model achieves 79.2\% AUC and 41.8\% MAP, indicating that the dataset presents a challenging problem for future research by the 
GIR community. 
The result technology can help
exploit the thematic content of web documents and social media for characterisation of locations.

\end{abstract}

%
%




\maketitle
\section{Introduction}
 
Location-based services (LBS) and the underlying Point of Interest (POI) datasets play a increasingly important role in our daily interaction with mobile devices. 
Platforms such as Yelp,
Foursquare, Google Map
allow users to search nearby POIs based on their names, place types, or tags, 
which requires manual data annotation. 
In fact, besides these structured data,  POIs are typically associated with abundant unstructured data such as descriptions and users' reviews which contain  useful information for search and question answering purpose. 
For example questions like ``Is this restaurant dog friendly?'' or ``Is this night club 18+?'' can be answered by relevant text in reviews such as ``Great dog friendly restaurant'' or ``18+ night club''. 
This information can also help accomplishing search needs such as ``find night clubs near me which are 18+''.


There are only a few
existing GIR benchmark datasets (e.g., GeoCLEF \cite{mandl2008geoclef}) 
and they often lack in rich annotations as would be required for 
the examples above. Recently many datasets have been produced for reading comprehension 
such as SQuAD \cite{rajpurkar2016squad}.
However, they do not have a spatial/platial component. Here we present a POI search and question answering  dataset called {\dataset} with detail annotations of context and answers.
Baseline models are implemented to demonstrate the difficulty of this task. 

Our work provides an evaluation benchmark for geographic information retrieval and question answering systems. It follows the idea of semantic signatures for social sensing ~\cite{usss} by which we can study POI types using patterns extracted from human behavior, e.g., what people write about places of a particular type. Intuitively, questions about age limits only arise in the narrow context of a few such types, e.g., nightclubs, movie theaters, and so on. Furthermore, unstructured data such as reviews are often geo-indicative without the need for explicit geographic coordinates. For instance, people may be searching for a \textit{central but quiet hotel} \cite{jwl}. It is those questions that \mbox{we will address in the following.}

\section{The {\dataset} Task}
We created {\dataset} based on the Yelp Challenge 11 (YC11) dataset\footnote{\url{https://www.yelp.com/dataset/challenge} 
} and the QA section of POI pages.

\paragraph{Query Set Generation}
We create the question answer dataset from the ``Ask the Community'' section\footnote{\url{https://www.yelpblog.com/2017/02/qa}} of POI pages.
The Yelp platform is dominated by popular business types such as restaurants. In order to produce a balanced query set for all business types we performed stratified sampling:
1) count the frequencies of POI name suffixes (single words) in YC11;
2) for every suffix with at least frequency 10 we create a quoted search query restricting to the Yelp business QA  domain\footnote{
Search ``site:https://www.yelp.com/questions/ `Restaurant" via Google}, and collect community QA page URLs from Google search engine;
3) collect questions and answers from the community QA pages.
In total, 1,701 quoted search queries results are collected from Google with up to 100 search results for each query. Since the last term often indicates the place type of a POI, the collected Yelp business question pages have a wide coverage of different place types. In total 20K questions were collected from Yelp business question pages for 1022 Yelp business types. Each question is associated with one or multiple POIs with several POI types, e.g.,  \textit{Echoplex} (\texttt{Music Venues}, \texttt{Bars}, \texttt{Dance Clubs}) or \textit{Paper Tiger Bar} (\texttt{Cocktail Bars}, \texttt{Lounges}). 

\paragraph{Relevance and Answer Annotation}
For each question, 10 review candidates are selected by stratified sampling from the search result of a lucene-based setup, i.e., applying \textit{Elastic Search} to POI reviews based on the question with constraint to the associated POI types. 
We developed a crowd-facing Web server and deployed it on Amazon Mechanical Turk to let raters annotate each sentence of these 10 reviews with respect to whether it answer the current question and what the corresponding answer is. 
The annotation results are collected for each question. To date, we have collected about 4100 questions. Basic statistic for these are shown in Tab. \ref{tab:stat}. In order to study the relationship between raters (given 3 raters per review sentence) and the accuracy of the raters, we divide the sentences into 4 sets based on the number of raters that agreed on each sentence, denoted as $R_{0}$, $R_{1}$, $R_{2}$, $R_{3}$. Then we randomly sample 20 sentences from each of the last three sets ($R_{1}$, $R_{2}$, $R_{3}$). By manually inspecting the relevance of these sentences to the corresponding questions. The resulting accuracy of each sample set is 45\% for $R_{1}$, i.e., 9/20 sentences, 90\% for $R_{2}$, 100\% for $R_{3}$. We treat the sentences in $R_{2}$, $R_{3}$ as relevant, and 
the rest are labeled as irrelevant sentences. These labels are used to evaluate different models.

\paragraph{Evaluation Metrics}
Area under curve (AUC) and mean average percision (MAP) are used as evaluation metrics.


{\small \begin{table}[]
\centering
	\caption{The Statistic of {\dataset}}
    \vspace{-0.05in}
    \label{tab:stat}
    \footnotesize
\begin{tabular}{l | r}
\toprule
\# of Annotated question                    & 4,100     \\ \hline
\% of questions WITHOUT related reviews & 11.4\%  \\ \hline
Avg. \# of related \textit{reviews} per question           & 4.61 \\ \hline
Avg. \# of 1 rater agreeing on relevant \textit{sentence} per question  & 2.19 \\ \hline
Avg. \# of 2  raters agreeing on relevant \textit{sentence} per question & 1.08 \\ \hline
Avg. \# of 3  raters agreeing on relevant \textit{sentence} per question & 0.83 \\ \bottomrule
\end{tabular}
\vspace{-0.15in}
\end{table}}

\begin{table}[h]
	\centering
	\caption{
        Examples of {\dataset}. 
       Each example consists of a question \textbf{Q}, one or more POI types \textbf{T},  a context sentence \textbf{C} from the POI reviews, and an answer \textbf{A}.      
       The ranking of sentence (C) based on human judgements (\textbf{H}), Lucene (\textbf{L}), and sentence embedding (\textbf{E}) is also shown. 
}
	\label{tab:qa_fail}
    \vspace{-0.05in}
	\small
	\begin{tabular}{p{1.2cm} | p{4.8cm} | p{1.1cm} 
    }
    \toprule
    
     Reason  & Example    & Ranking (H/L/E) \\
     \hline
\multirow{4}{1.3cm}{Paraphrase} 
& \textbf{Q:} About \textbf{how long} should I expect my visit to be?   
&\\
       & \textbf{T:} 
       Venues \& Event Spaces; Kids Activities       & 1/107/88             \\
       & \textbf{C:} We \textbf{were there for} about 2 hours, including the show.  &  out of 158           \\
       & \textbf{A:} took 2 hrs   &                 \\ \hline
\multirow{4}{1.3cm}{Hyponym} 
& \textbf{Q:} Any good \textbf{vegan choices}?  & 
\\
       & \textbf{T:} Restaurants$\rightarrow$Cajun/Creole & 2/49/18  \\
       & \textbf{Sent:} After scanning the menu for a bit however, I was able to find the \textbf{tofu} wings.   &  out of 83        \\
       & \textbf{A:} Tofu wings could be a choice                 &    \\ \hline
\multirow{4}{1.3cm}{Synonymy} 
& \textbf{Q:} Any recommendations on how to \textbf{score a table}? ...
&\\
       & \textbf{T:} Restaurants$\rightarrow$French            & 1/63/14 \\
       & \textbf{C:} I \textbf{made a reservation} a day in advance thinking it will be busy.                          &     out of 98  \\
       & \textbf{A:} A day in advance                       &      \\ \hline
\multirow{4}{1.3cm}{Deduction} & \textbf{Q:} Are there classes for \textbf{seniors}?   
&\\
       & \textbf{T:} 
        Art Galleries; Art Schools        &  1/60/15 \\
       & \textbf{C:} Great studio for \textbf{all}, including kids!   &    out of 72  \\
       & \textbf{A:} There are classes for seniors    &    \\ \hline
\multirow{4}{1.3cm}{
Common Sense }
& \textbf{Q:} Do they \textbf{buy comic books}?   
&\\
       & \textbf{T:} Shopping$\rightarrow$Comic Books          & 1/45/53\\
       & \textbf{C:} The concerns:  The store currently has \textbf{no consignment  or new issues}.       &  out of 62  \\
       & \textbf{A:} No    & \\ 
\bottomrule
\end{tabular}
\vspace{-0.2in}
\end{table}

\section{Experiment with Baseline Models} \label{sec:baseline}
In order to provide a similar search functionality to Yelp's new 
{review-based POI search}\footnote{https://engineeringblog.yelp.com/2017/06/moving-yelps-core-business-search-to-elasticsearch.html}, we developed a \textit{TF-IDF based model}  to search through all sentences from 10 reviews based on a question.
An evaluation using the {\dataset} dataset gives 77\% AUC and 48.8\% MAP. 
%
We also applied
the \textit{sentence embedding model} proposed by Sanjeev Arora et al.~ \cite{arora2016simple}. It improves the average word embeddings 
using SVD and gives what the authors call ``tough-to-beat'' results for a text similarity tasks.
We use the pretrained Google News 300 dimension Word2Vec embeddings to generate the sentence level embedding for both questions and review sentences. Then their cosine similarities
are used to rank the sentences given a question. 
Evaluation by {\dataset} gives 79.2\% AUC and 41.0\% MAP. Comparing to the TF-IDF model, the sentence embedding-based model gives a higher AUC (which is sensitive to overall rankings) but lower MAP (which is sensitive to  top rankings). 
The results from both baseline models indicate that the {\dataset} dataset presents a challenging task. Table \ref{tab:qa_fail} shows examples for which the baseline model fails. 
Correctly predicting relevant sentence requires an understanding of language and common sense. We hope that the dataset will enable further GIR research about question answering as it relates to place types.

\bibliographystyle{ACM-Reference-Format}
\bibliography{sample-bibliography}

\end{document}